\documentclass[conference]{IEEEtran}
\usepackage{cite}

\usepackage{amsmath,amssymb,amsfonts}
\usepackage{algorithmic}
\usepackage{graphicx}
\usepackage{textcomp}
\usepackage[colorlinks,backref=page]{hyperref}
\usepackage{booktabs}
\usepackage{caption}
\usepackage{subcaption}
\usepackage{xcolor}
\usepackage{amssymb}
\usepackage{balance}

\def\BibTeX{\rm B\kern-.05em{\sc i\kern-.025em b}\kern-.08em
    T\kern-.1667em\lower.7ex\hbox{E}\kern-.125emX}

\makeatletter

\def\ps@IEEEtitlepagestyle{%
  \def\@oddfoot{\mycopyrightnotice}%
  \def\@evenfoot{}%
}
\def\mycopyrightnotice{%
  {\footnotesize Preprint, Peer reviewed \& Accepted @ EUVIP 2023\hfill}
  \gdef\mycopyrightnotice{}
}
\begin{document}
\title{MAiVAR-T: Multimodal Audio-image and Video Action Recognizer using
Transformers}

\author{\IEEEauthorblockN{Muhammad Bilal Shaikh\IEEEauthorrefmark{1},Douglas Chai\IEEEauthorrefmark{2}}
\IEEEauthorblockA{School of Engineering\\
Edith Cowan University\\
\IEEEauthorrefmark{1}mbshaikh@our.ecu.edu.au, \IEEEauthorrefmark{2}d.chai@ecu.edu.au}
\and
\IEEEauthorblockN{Syed Mohammed Shamsul Islam}
\IEEEauthorblockA{School of Science\\
Edith Cowan University\\
syed.islam@ecu.edu.au}
\and
\IEEEauthorblockN{Naveed Akhtar}
\IEEEauthorblockA{Department of Computer Science \& Software Engineering\\
The University of Western Australia\\
naveed.akhtar@uwa.edu.au}}

\maketitle
\begin{abstract} In line with the human capacity to perceive the world by simultaneously processing and integrating high-dimensional inputs from multiple modalities like vision and audio, we propose a novel model, MAiVAR-T (Multimodal Audio-Image to Video Action Recognition Transformer). This model employs an intuitive approach for the combination of audio-image and video modalities, with a primary aim to escalate the effectiveness of multimodal human action recognition (MHAR). At the core of MAiVAR-T lies the significance of distilling substantial representations from the audio modality and transmuting these into the image domain. Subsequently, this audio-image depiction is fused with the video modality to formulate a unified representation. This concerted approach strives to exploit the contextual richness inherent in both audio and video modalities, thereby promoting action recognition. In contrast to existing state-of-the-art strategies that focus solely on audio or video modalities, MAiVAR-T demonstrates superior performance. Our extensive empirical evaluations conducted on a benchmark action recognition dataset corroborate the model's remarkable performance. This underscores the potential enhancements derived from integrating audio and video modalities for action recognition purposes.\end{abstract}

\section{Introduction}

Human action recognition has become a critical task in various fields such as surveillance \cite{park2021skeletonvis}, robotics \cite{scirobo}, interactive gaming \cite{oinas2018persuasive}, and health care \cite{liu2022overview}. Traditionally, most approaches have focused on visual cues \cite{li2019collaborative}. However, human actions are not limited to visual manifestations; they also consist of rich auditory information \cite{gao2020listentolook}. Accordingly, Multimodal human action recognition (MHAR) that incorporates both visual and audio cues can provide more comprehensive and accurate recognition results \cite{shaikh2021rgb}. \par

Despite these promising prospects, the performance of current MHAR models is hampered by challenges of multimodal data fusion. Existing methods, including Convolutional Neural Networks (CNNs) \cite{lecun1989cnn,krizhevsky2017imagenet,he2016deep} require significantly more computation than their image counterparts, some architectures factorise convolutions across spatiotemporal dimensions. Contrastingly, Recurrent Neural Networks (RNNs) and Long Short-Term Memory (LSTMs) \cite{hochreiter1997long} have demonstrated constraints in processing large sequences, memory efficiency and parallelism. \par

 In this paper, we propose a novel transformer-based model, Multimodal Audio-image and Video Action Recognizer using Transformers (MAiVAR-T). Our approach capitalizes on the self-attention mechanism inherent in transformers \cite{vaswani2017attention} to extract relevant features from both modalities and fuse them effectively. The proposed MAiVAR-T model outperforms state-of-the-art MHAR models on benchmark datasets \cite{soomro2012ucf101}, demonstrating the potential of transformer-based architectures in improving multimodal fusion and recognition accuracy. \par
\begin{figure*}
    \centering
    \includegraphics[width=0.98\textwidth]{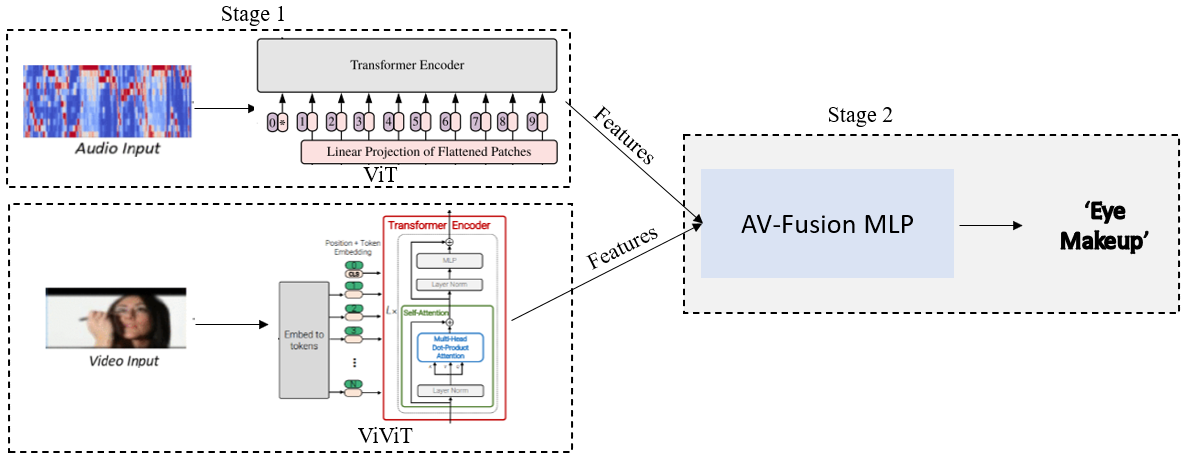}
    \caption{The proposed framework contains two stages. The first stage extracts the features influencing the recognition while the second stage performs classification on the fused features. The input sequence
consists of image and audio-image patches. These are then projected into tokens and appended to
special CLS (classification). Our transformer encoder then uses
self attention to model unimodal information, and send cross-modal information flow through to fusion network.}
    \label{fig:framework}
\end{figure*}

 To summarize, the contributions made in this paper are:
\begin{itemize}
    \item   A new feature representation strategy is proposed to select the most informative candidate representations for audio-visual fusion; 
    \item Collection of effective audio-image-based representations that complement video modality for better action recognition are included; 
    \item We apply a novel MAiVAR-T framework (see Fig. \ref{fig:framework}) for audio-visual fusion that supports different audio-image representations and can be applied to different tasks; and 
    \item State-of-the-art results for action recognition on the audio-visual dataset have been reported.
\end{itemize}
The remainder of the paper is organized as follows: we begin with a review of related works on MHAR  (Section \ref{related}), followed by a detailed discussion of the proposed methodology (Section \ref{method}). We then present the experimental setup (Section \ref{evaluation}) and report the results (Section \ref{results}). Finally, 
we concludes the paper with future directions (Section \ref{conclusion}).

\section{Related Work}\label{related}

\subsection{Deep Learning for MHAR}

Recently, deep learning models have shown remarkable results in MHAR \cite{shaikh2022maivar}. They are capable of automatically learning a hierarchy of intricate features from raw multimodal data, which are beneficial for action recognition tasks. 

CNNs have been widely adopted for MHAR to automatically extract spatial features from input data \cite{lin2019tsm}, and LSTMs are typically used for modelling the temporal dynamics of actions \cite{hochreiter1997long}. However, the traditional combination of CNNs and LSTMs for MHAR faces challenges such as ineffective multimodal fusion and difficulty handling long temporal sequences.

Transformers, introduced by Vaswani et al. \cite{vaswani2017attention}, have demonstrated their superiority in many fields like natural language processing \cite{devlin2018bert}, image classification \cite{dosovitskiy2020image}, and video understanding \cite{Arnab_2021_ICCV}. The self-attention mechanism through its optimal complexity (see Table \ref{complexity}) in transformers could potentially enhance the capability of feature extraction and multimodal fusion in MHAR tasks. However, the utilization of transformers in MHAR is relatively unexplored and demands further investigation. 

\subsection{Audiovisual Learning and Fusion}

The field of audiovisual multimodal learning has a long and diverse history, both preceding and during the deep learning era \cite{deepmm2017rama}. Early research focused on simpler approaches, utilizing hand-designed features and late-stage processing, due to limitations in available data and computational resources \cite{av1998chen}. However, with the advent of deep learning, more sophisticated strategies have emerged, enabling the implicit learning of modality-specific or joint latents to facilitate fusion. As a result, significant advancements have been achieved in various supervised audiovisual tasks \cite{deep2013kim}.

It is common to jointly train multiple modality-specific convolution networks, where the intermediate activations are combined either through summation \cite{kazakos2019epicfusion}. On the other hand, in transformer-based architectures, the incorporation of Vision Transformers (ViT) \cite{dosovitskiy2020image} and Video Vision Transformers (ViViT) \cite{Arnab_2021_ICCV} has brought about significant advancements in multimodal human action recognition. Initially, ViT proved instrumental in dissecting images into smaller segments, to interpret these patches as a sequence for more accurate image understanding. This ability greatly improved the recognition and classification of human actions within still images. The introduction of ViViT further extended this capacity, applying transformer techniques to analyze video data. By processing sequences of video frames, ViViT effectively interprets the spatio-temporal dynamics involved in human movements. Together, the use of Vision Transformers and Video Vision Transformers can produce a shift in multimodal human action recognition, enhancing the capability of systems to accurately classify and understand complex human activities across visual and audio domains.

\begin{table}
        \caption{Complexity comparison for different types of layer. Notations: $n$ : sequence length, $d$: representation dimension, $k$ kernel size.}
\begin{center}
\begin{tabular}{|l|p{15mm}|p{15mm}|p{15mm}|}
\hline
Layer Type        & Complexity per layer &  Sequential Operations & Maximum Path Length   \\[0.5ex] 
\hline
Convolutional & $O(k\cdot n \cdot d^2)$ & $O(1)$ & $O(log_k (n)) $   \\ 
Recurrent &  $O(n\cdot d^2)$ & $O(n)$ & $O(n)$  \\
Self-Attention & $ O(n^2 \cdot d ) $ & $ O(1$ & $O(1)$ \\ 
\hline
\end{tabular}
\end{center}
    \label{complexity}
\end{table}

\begin{table}[]
    \centering
    \caption{Hyper-parameters of the network.}
    \begin{tabular}{|l|c|}
    \hline
        Parameter & Value  \\
        \hline
Batch size & 256 \\
Initial learning rate &  0.001 \\
lr decay (every 4 epochs) &  0.10 \\
Learning rate patience &  10 \\ 
Epochs &  100 \\
\hline

    \end{tabular}
    
    \label{tab:hparams}
\end{table}

\section{Proposed Methodology}\label{method}

\textbf{Data Collection:}
 We collected human actions from a benchmark dataset called UCF101 \cite{soomro2012ucf101}, with each instance containing video clips and their corresponding audio streams. UCF-101 contains an average length of $180$ frames per video. We observed that half of the videos in the dataset contained no audio. Thus, in order to focus on the effect of audio features, we used only those videos that contained audio. This resulted in $6837$ videos across $51$ categories. Whilst this led the dataset to be significantly reduced,  the distribution of the audio dataset was similar to the video dataset. We used the first train-test split setting provided with this dataset,  which resulted in $4893$ training and $1944$ testing samples. We reported the top 1 accuracies obtained by training on split $1$. 

\textbf{Data Preprocessing:}
The video and audio data were preprocessed separately, as described in the following subsections. The video data was transformed into frames, while the audio data was converted into six audio-image representations following \cite{shaikh2023pymaivar,shaikh2022maivar}. Standard normalization techniques were applied to both modalities. 

\begin{figure} 
    \centering
    
    \begin{subfigure}[b]{0.85\linewidth}
         \centering
         \includegraphics[width=\textwidth]{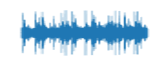}
    \caption{Waveplot}
         
     \end{subfigure}\hfil
     \begin{subfigure}[b]{0.85\linewidth}
         \centering
         \includegraphics[width=\textwidth]{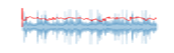}
    \caption{Spectral Centroids}
         
     \end{subfigure}\hfil
     \begin{subfigure}[b]{0.85\linewidth}
         \centering
         \includegraphics[width=\textwidth]{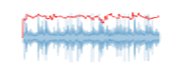}
    \caption{Spectral Rolloff}
         
     \end{subfigure}

     \begin{subfigure}[b]{0.85\linewidth}
         \centering
         \includegraphics[width=\textwidth]{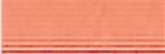}
    \caption{MFCCs}
         
     \end{subfigure}\hfil
     \begin{subfigure}[b]{0.85\linewidth}
         \centering
         \includegraphics[width=\textwidth]{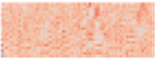}
    \caption{MFCCs Feature Scaling}
         
     \end{subfigure}\hfil
     \begin{subfigure}[b]{0.85\linewidth}
         \centering
         \includegraphics[width=\textwidth]{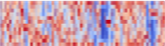}
    \caption{Chromagram}
         
     \end{subfigure}\hfil
     \begin{subfigure}[b]{0.85\linewidth}
         \centering
         \includegraphics[width=\textwidth]{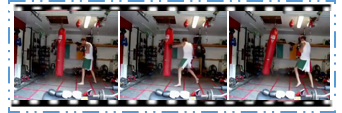}
    \caption{Video input}
         
     \end{subfigure}
    \caption{Segmented video input and six different audio-image representations of the same action.}
    
    \label{fig:input-reps}
\end{figure}

\textbf{Audio image representations:} Following are some of the key characteristics of audio-image representations (shown in Figure \ref{fig:input-reps}).
\begin{itemize}
    \item Audio image representations provide a significant reduction in dimensionality. For example, spectral centroid images represent the frequency content of the audio signal over time, which is a lower-dimensional representation of the original video dataset. This can make it easier and faster to process the data and extract meaningful features.
    \item Audio images are based on the audio signal, which is less affected by visual changes, such as changes in lighting conditions or camera angles. This makes these representations more robust to visual changes and can improve the accuracy of human action analysis.
    \item Standardization as audio images can be standardized to a fixed size and format, which can make it easier to compare and combine data from diverse sources. This can be useful for tasks such as cross-dataset validation and transfer learning. Hence, this dataset can serve as a standard benchmark for evaluating the performance of different machine-learning algorithms for human action analysis based on audio signals.
    \item Suitable for privacy-oriented applications such as surveillance or healthcare monitoring, which may require the analysis of human actions without capturing the original visual information. 
\end{itemize}

\begin{figure}
     \centering
     \begin{subfigure}[b]{0.23\textwidth}
         \centering
         \includegraphics[width=\textwidth]{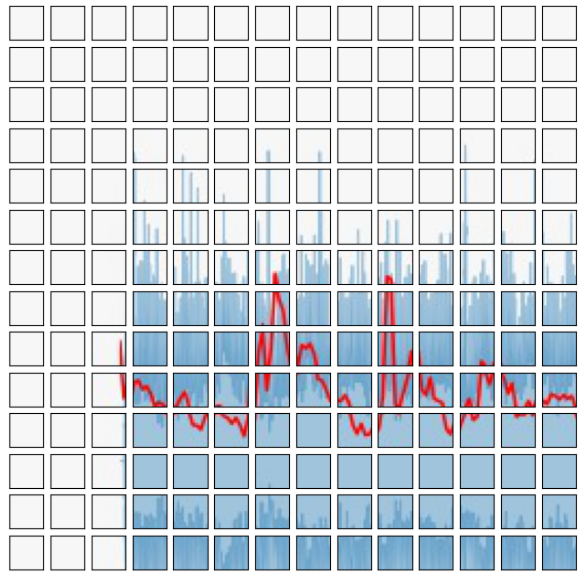}
         \caption{}
         \label{fig:a}
     \end{subfigure}
     \hfill
     \begin{subfigure}[b]{0.23\textwidth}
         \centering
         \includegraphics[width=\textwidth]{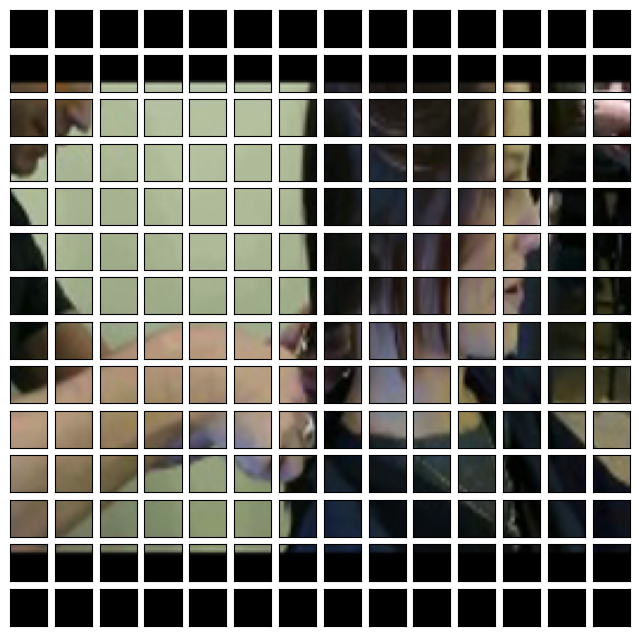}
         \caption{}
         \label{fig:b}
     \end{subfigure}
        \caption{Image patches (a) Audio-image representation, (b) RGB video frame.}
        \label{fig:pacthes}
\end{figure}


\begin{figure}
    \centering
    \includegraphics[width=0.4\textwidth,scale=0.6]{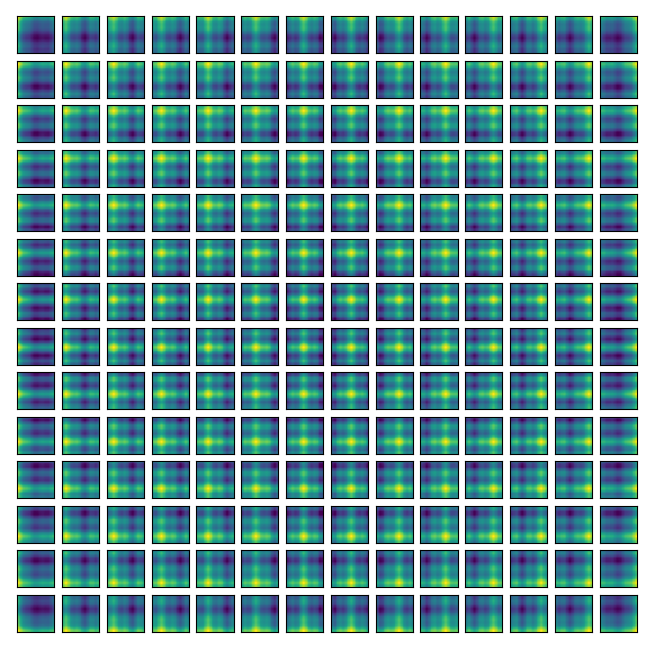}
    \caption{Positional embeddings.}
    \label{fig:pos-embed}
\end{figure}

\textbf{Architecture:} The MAiVAR-T model comprises an audio transformer, a video transformer, and a cross-modal attention layer. The transformers process the audio and video inputs separately, after which the cross-modal attention layer fuses the outputs. Finally, a classification layer predicts the action present in the input data.

\textbf{Audio Stream:} The audio stream uses Vision Transformer (ViT) \cite{dosovitskiy2021an} to process 2D images with minimal
changes. In particular, ViT extracts $N$ non-overlapping image patches, $x_i  \in \mathbb{R}^{h\times w}$, performs a linear projection and
then rasterises them into 1D tokens $z_i \in \mathbb{R}^d$. The sequence
of tokens input to the following transformer encoder is

\begin{equation}
\textbf{z} = [z_{cls}, \textbf{E}x_1, \textbf{E}x_2, . . . , Ex_N ] + \textbf{p},     
\end{equation}
where the projection by E is equivalent to a 2D convolution. In addition, a
learned positional embedding, $p \in \mathbb{R}^
{N\times d}$
, is added to the
tokens to retain positional information, as the subsequent
self-attention operations in the transformer are permutation
invariant. The tokens are then passed through an encoder
consisting of a sequence of $L$ transformer layers. The MLP consists of two linear projections separated by a
GELU non-linearity and the token-dimensionality, $d$,
remains fixed throughout all layers. Finally, a linear classifier is used to classify the encoded input based on $z^{L}_{cls} \in \mathbb{R}^d$,
if it was prepended to the input, or a global average pooling
of all the tokens, $z^L$, otherwise.
As the transformer \cite{vaswani2017attention}, which forms the basis of
ViT \cite{dosovitskiy2020image}, is a flexible architecture that can operate on any
sequence of input tokens $z \in \mathbb{R}^{N \times d}$, we describe strategies
for tokenising videos next.

\textbf{Video Feature Stream:}
We consider mapping a video $\mathbb{V} \in \mathbb{R}^{T \times H \times W \times C}$ to a sequence of tokens $z' \in
\mathbb{R}^{n_t \times n_h \times n_w \times d}$. We then add the positional embedding and
reshape into $\mathbb{R}^{N \times d}$
to obtain $z$, the input to the transformer.

\section{Experiments}\label{evaluation}

\subsection{Audio preprocessing}
\begin{figure*}
    \centering
    \includegraphics[width=0.85\textwidth]{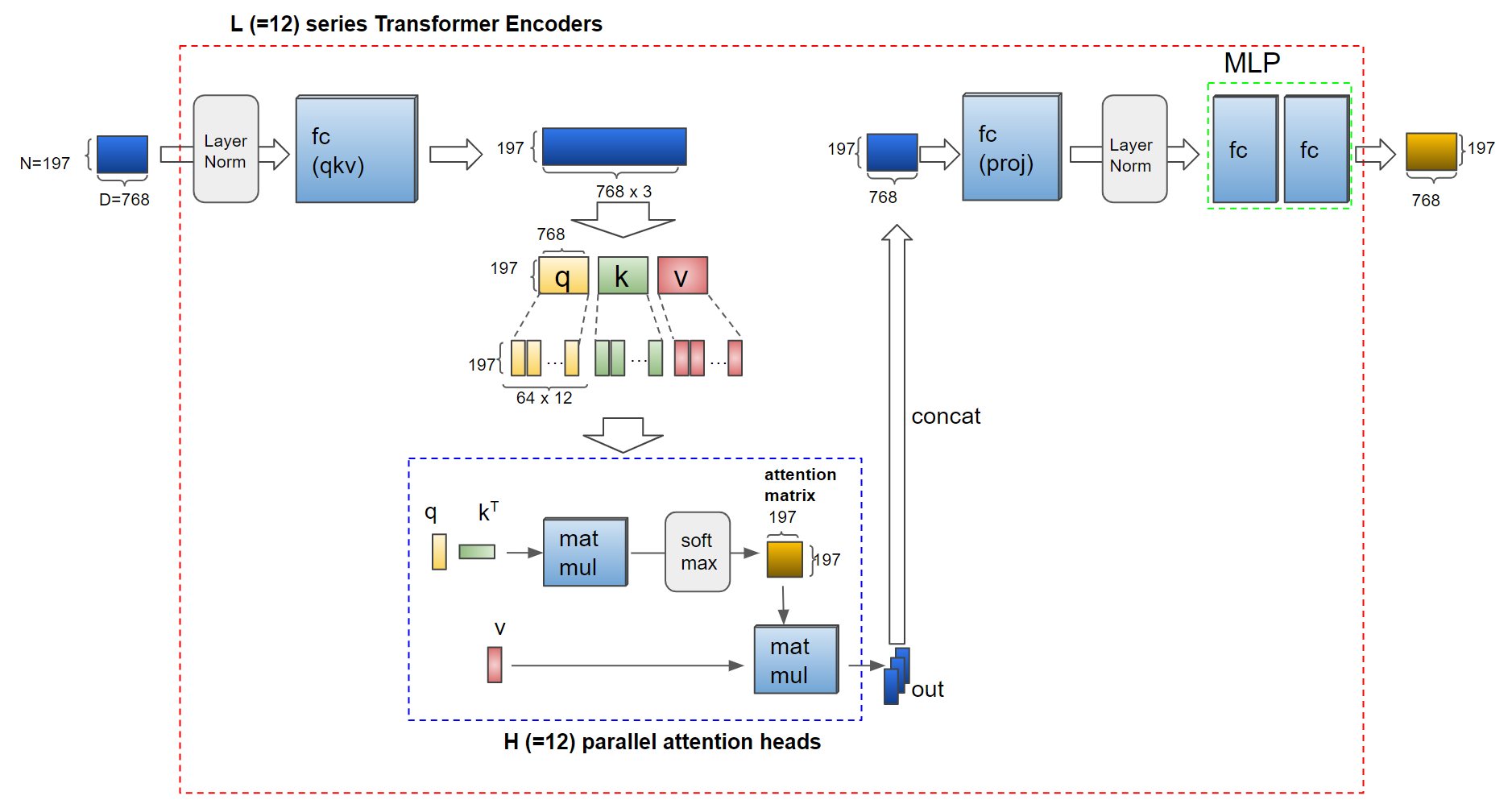}
    \caption{Schematic of Vision Transformer Encoder.}
    \label{schematic}
\end{figure*}

Each audio image representation was broken into patches as illustrated in the examples shown in Figure \ref{fig:pacthes}. For spatial context, positional embeddings for each input were projected into the architecture (see Figure \ref{fig:pos-embed}). An internal schematic of the transformer model has been illustrated in Figure \ref{schematic}. Training data was batched into mini-batches of 16 instances each. Augmentation techniques like random cropping and time-stretching were applied to increase model robustness.

\subsection{Video preprocessing}

Following \cite{Arnab_2021_ICCV}, the features extracted are then fed to the multimodal fusion module (AV-Fusion MLP) which later performs the classification for each action class.

\subsection{Training}

We utilized a multimodal cross-entropy loss function for training, balancing both audio and video modalities. The network hyperparameters are reported
in Table \ref{tab:hparams}.

\textbf{Hardware and Schedule:}
The training was performed on a high-performance computing cluster, equipped with GeForce GTX 1080 Ti GPUs. We trained the transformer-based model for 100 epochs, with a learning rate ($ \alpha $) schedule that decreased the rate by 10\% every 4 epochs.
\textbf{Optimizer:} The Adam optimizer \cite{kingma2014adam} was used due to its effectiveness in training deep networks.
\textbf{Regularization:} Dropout techniques \cite{srivastava2014dropout} were applied to prevent overfitting during training.

\section{Results}\label{results}

\begin{figure}[t]
    \centering    \includegraphics[width=0.45\textwidth,scale=0.8]{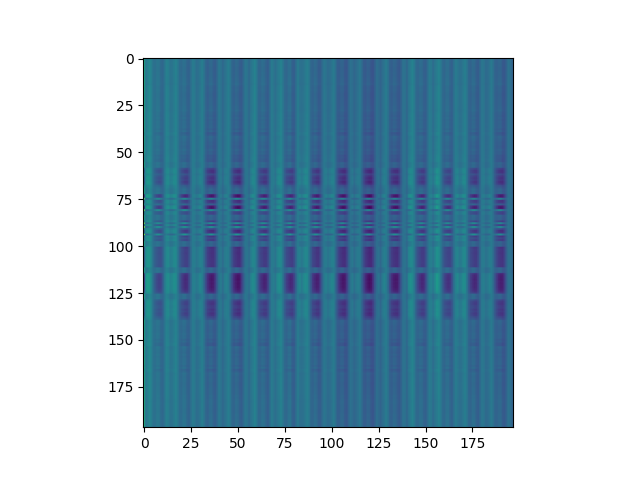}
    \caption{Attention matrix for an audio-image representation.}
    \label{fig:attention}
\end{figure}
\begin{figure}   
    \centering    \includegraphics[width=0.48\textwidth]{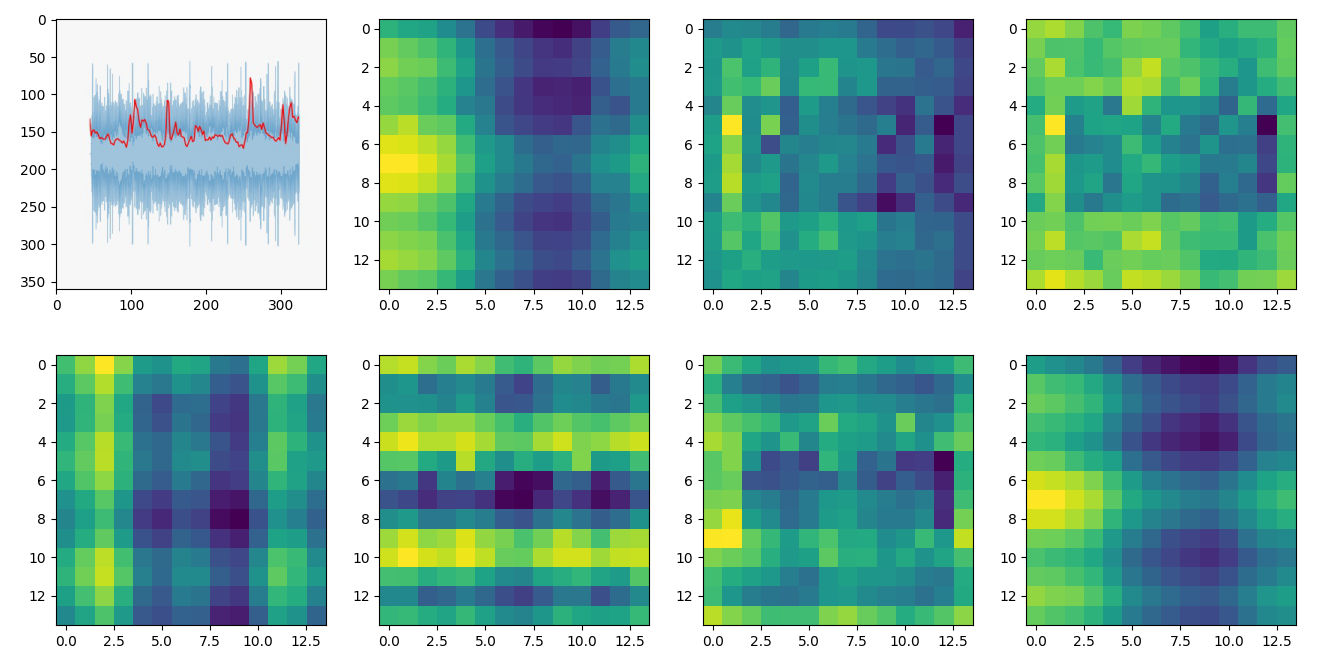}
    \caption{Visualization of attention.}
    \label{fig:attention-flow}
\end{figure}
To assess the contribution of each component in our model, we performed an ablation study. Results demonstrate that both the audio and video transformers, as well as the cross-modal attention layer, contribute significantly to the final action recognition performance. The process of attention mechanism in the extraction of features through robust audio-image representations could be visualized in Figures \ref{fig:attention} and \ref{fig:attention-flow}. We have used an accuracy metric that measures the proportion of correct predictions made by the model out of all the predictions and defined as:
\begin{equation}
\text{Accuracy} = \frac{TP + TN}{TP + FP + TN + FN}
\end{equation}
\noindent where $TP$ are the correctly predicted positive values. $TN$ are the correctly predicted negative values. $FP$, also known as Type I errors, are the negative values incorrectly predicted as positive. $FN$, also known as Type II errors, are the positive values incorrectly predicted as negative.

Table \ref{audio-ablation} compares the performance of transformer-based feature extractors with CNN-based counterparts. Proposed MAiVAR-T outperforms prior methods by a +3\% as presented in Table \ref{sota-table}.

\begin{table}[t]
    \centering
    \caption{Test accuracy of different audio representations with CNN and transformer-based backbones (InceptionResNet-v4(IRV4) and Vision Transformer (ViT) respectively)}
\begin{center}
\begin{tabular}{|c|c|c|}
\hline 
\textbf{Representation}           & \textbf{IRV4} & \textbf{ViT}  \\
\hline
Waveplot                 & 12.08 &  \textbf{19.7 (+7)} \\
Spectral Centroids        & 13.22  & \textbf{28.65 (+15) }        \\   
Spectral Rolloff              & 16.46 &\textbf{ 26.85 (+10) }     \\
MFCCs                & 12.96  &  \textbf{18.26 (+6)}    \\
MFCCs Feature Scaling                & 17.43  & \textbf{17.44 (+0.01)}    \\
Chromagram                        & 15.48 & \textbf{19.08 (+3)} \\
\hline
\end{tabular}
\end{center}    
    \label{audio-ablation}
\end{table}

\begin{table}[t]
    \centering
    \caption{Classification accuracy of MAiVAR compared to the state-of-the-art methods on UCF51 dataset after fusion of audio and video features.}
\begin{center}
\begin{tabular}{|c| c| c|}
\hline
\textbf{YEAR} & \textbf{METHOD}        & \textbf{ACCURACY} [\%] \\[0.5ex] 
\hline

2015 &  C3D \cite{tran2015learning}                      & 82.23              \\
2016 & TSN (RGB) \cite{wang2016temporal}                & 60.77    \\
2017 & C3D+AENet \cite{takahashi2017aenet}                 & 85.33             \\
2018 & DMRN\cite{tian2018audio}                 & 81.04    \\
2018 &  DMRN\cite{tian2018audio}  +\cite{brousmiche2021multi} features                 & 82.93     \\
2020 &  Attention Cluster \cite{long2020purely}                 & 84.79     \\
 2020 & IMGAUD2VID \cite{gao2020listentolook}                & 81.10 \\
2022 & STA-TSN (RGB)\cite{yang2022sta} & 82.1      \\
2022 &  MAFnet \cite{brousmiche2021multi}                 & 86.72   \\

\hline
2022 &  MAiVAR-WP\cite{shaikh2022maivar}             & 86.21   \\
2022  & MAiVAR-SC\cite{shaikh2022maivar}            & 86.26   \\
2022  & MAiVAR-SR\cite{shaikh2022maivar}             & 86.00    \\
2022  & MAiVAR-MFCC \cite{shaikh2022maivar}            & 83.95    \\
2022  &  MAiVAR-MFS\cite{shaikh2022maivar}             & 86.11    \\
2022  &  MAiVAR-CH \cite{shaikh2022maivar}            & 87.91  \\
\hline
\textbf{Ours} &  \textbf{MAiVAR-T}          &  \textbf{91.2} \\

\hline
\end{tabular}
\end{center}
    \label{sota-table}
\end{table}

\section{Conclusion}\label{conclusion}
Over the past decade, Convolutional Neural Networks (CNNs) with video-based modalities have been a staple in the field of action video classification. However, in this paper, we challenge the indispensability of video modalities and propose a transformer-based multi-modal audio-image to video action recognition framework called Multi-modal Audioimage-Video Action Recognizer using Transformers (MAiVAR-T). This fusion-based, end-to-end model for audio-video classification features a transformer-based architecture that not only simplifies the model but also enhances its performance.

Experimental results demonstrate that our transformer-based audio-image to video fusion methods hold their own against traditional image-only methods, as corroborated by previous research. Given the significant improvements observed with pre-training on larger video datasets, there is considerable potential for further enhancing our model's performance.
In future work, we aim to validate the efficacy of integrating text modality with audio and visual modalities. Furthermore, the scalability of MAiVAR-T on large-scale audio-video action recognition datasets, such as Kinetics $400/600/700$ will be explored. Additionally, we plan to explore better architectural designs to integrate our proposed approach with more innovative ideas, such as integrating generative AI-based transformer architectures, into our network could provide valuable insights into the impact of transformers on MHAR.

 \section*{Acknowledgment}
 This work is jointly supported by Edith Cowan University (ECU) and the Higher Education Commission (HEC) of Pakistan under Project \#PM/HRDI-UESTPs/UETs-I/Phase-1/Batch-VI/2018. Dr. Akhtar is a recipient of Office of National Intelligence National Intelligence Postdoctoral Grant \# NIPG-2021–001 funded by the Australian Government.





\balance
\bibliographystyle{IEEEtran}
\bibliography{IEEEabrv,main}

\end{document}